\documentclass[11pt,a4paper]{article}

\usepackage[utf8]{inputenc}
\usepackage{amsmath}
\usepackage{amsfonts}
\usepackage{amssymb}
\usepackage{graphicx}
\usepackage{authblk}
\usepackage[margin=1in]{geometry}
\usepackage{hyperref}

\title{\textbf{Enhancing Hyperspace Analogue to Language (HAL) Representations via Attention-Based Pooling for Text Classification}}

\author[1]{Ali Sakour\thanks{Corresponding Author. Email: alisakour78@gmail.com}}
\author[1]{Zoalfekar Sakour}

\affil[1]{Undergraduate Students, Department of Computer and Automatic Control Engineering, \authorcr 
Faculty of Mechanical and Electrical Engineering, \authorcr
Lattakia University, Lattakia, Syria}

\date{}

\begin{document}

\maketitle

\begin{abstract}
The Hyperspace Analogue to Language (HAL) model relies on global word co-occurrence matrices to construct distributional semantic representations. While these representations capture lexical relationships effectively, aggregating them into sentence-level embeddings via standard mean pooling often results in information loss. Mean pooling assigns equal weight to all tokens, thereby diluting the impact of contextually salient words with uninformative structural tokens. In this paper, we address this limitation by integrating a learnable, temperature-scaled additive attention mechanism into the HAL representation pipeline. To mitigate the sparsity and high dimensionality of the raw co-occurrence matrices, we apply Truncated Singular Value Decomposition (SVD) to project the vectors into a dense latent space prior to the attention layer. We evaluate the proposed architecture on the IMDB sentiment analysis dataset. Empirical results demonstrate that the attention-based pooling approach achieves a test accuracy of 82.38\%, yielding an absolute improvement of 6.74 percentage points over the traditional mean pooling baseline (75.64\%). Furthermore, qualitative analysis of the attention weights indicates that the mechanism successfully suppresses stop-words and selectively attends to sentiment-bearing tokens, improving both classification performance and model interpretability.
\end{abstract}

\vspace{1em}
\section{Introduction}

Distributional semantic models operate on the premise that words occurring in similar contexts possess similar meanings. The Hyperspace Analogue to Language (HAL) model \cite{lund1996producing} operationalizes this concept by constructing a high-dimensional word co-occurrence matrix based on a sliding window. By inversely weighting co-occurrences according to the distance between target and context words, HAL generates robust lexical representations that capture both syntactic and semantic word properties.

However, deploying word-level representations for sequence-level tasks, such as document classification or sentiment analysis, necessitates a vector aggregation strategy. The standard historical approach, mean pooling, computes the unweighted average of all constituent word vectors in a sequence. This uniform weighting assumes that all tokens contribute equally to the overall meaning of the text. Consequently, the semantic signal of highly discriminative tokens (e.g., sentiment-bearing adjectives) is systematically diluted by the inclusion of high-frequency, uninformative structural tokens (stop-words). 

To address this limitation, we propose an architecture that integrates classical distributional semantics with a learnable attention mechanism. Prior to aggregation, the raw, sparse HAL co-occurrence matrix---which scales linearly with the vocabulary size ($2V$)---is projected into a lower-dimensional dense latent space using Truncated Singular Value Decomposition (SVD). This step resolves the curse of dimensionality and reduces the computational overhead required for neural network training. 

Following dimensionality reduction, we replace standard mean pooling with a parameterized additive attention layer \cite{bahdanau2014neural}. Instead of a static average, the network learns to dynamically assign scalar weights to each word representation based on its relevance to the classification objective. To regulate the attention distribution and prevent the network from overfitting to single trigger words, we incorporate a temperature scaling factor into the softmax function.

The primary contributions of this paper are as follows:
\begin{itemize}
    \item We introduce a hybrid architecture that enhances static HAL representations by replacing traditional mean pooling with a learnable, temperature-scaled attention mechanism.
    \item We detail a preprocessing pipeline utilizing Truncated SVD to compress high-dimensional, sparse co-occurrence matrices into dense embeddings suitable for end-to-end neural network training.
    \item We provide empirical validation on the IMDB sentiment analysis dataset, demonstrating that the attention-based pooling mechanism yields an absolute accuracy improvement of 6.74 percentage points over the mean pooling baseline.
    \item We conduct a qualitative analysis of the attention weights, confirming that the proposed mechanism successfully filters out structural noise and correctly identifies salient tokens in mixed-sentiment contexts.
\end{itemize}
\section{Related Work}

\subsection{Distributional Semantics and Co-occurrence Models}
Early approaches to semantic representation relied extensively on word co-occurrence statistics derived from large corpora. Latent Semantic Analysis (LSA) \cite{deerwester1990indexing} applies Singular Value Decomposition (SVD) to document-term matrices to extract latent semantic structures. In contrast, the Hyperspace Analogue to Language (HAL) \cite{lund1996producing} constructs a word-word co-occurrence matrix by passing a sliding window over the text, assigning weights inversely proportional to the distance between target and context words. While these count-based models effectively capture lexical semantics, their raw representations are highly dimensional and sparse. Subsequent predictive models, such as Word2Vec \cite{mikolov2013distributed} and GloVe \cite{pennington2014glove}, popularized dense, low-dimensional embeddings. However, classical distance-weighted co-occurrence matrices remain theoretically robust foundations for distributional semantics when appropriately compressed.

\subsection{Sentence Representation and Pooling Strategies}
Transforming word-level embeddings into fixed-length sequence representations requires an aggregation mechanism. Standard baselines, such as the fastText classifier \cite{joulin2017fasttext}, compute a simple unweighted average of word vectors. While computationally efficient, this linear aggregation is agnostic to word importance, assigning equal weight to discriminative terms and semantically vacant structural tokens. This uniform weighting restricts the model's capacity to focus on sentiment-critical triggers, a limitation our proposed attention mechanism explicitly addresses.
\subsection{Attention Mechanisms in Text Classification}
Attention mechanisms, originally formalized for sequence-to-sequence neural machine translation \cite{bahdanau2014neural}, mitigate the information bottleneck by allowing models to dynamically weight input tokens. This concept was subsequently adapted for sequence classification. For example, Yang et al. \cite{yang2016hierarchical} introduced Hierarchical Attention Networks (HAN), demonstrating that learnable attention layers significantly outperform static pooling operations by selectively focusing on informative words.

\subsection{The Current Gap}
Contemporary natural language processing architectures predominantly pair attention mechanisms with predictive embeddings or fully contextualized Transformer models \cite{vaswani2017attention}. The integration of parameterized attention pooling directly with classical, count-based co-occurrence matrices, such as HAL, remains underexplored. This paper addresses this specific gap by demonstrating that HAL representations, when subjected to dimensionality reduction via SVD, can be effectively and efficiently aggregated using a temperature-scaled additive attention mechanism.
\section{Methodology}

The proposed architecture bridges static distributional semantics with dynamic neural aggregation. The pipeline comprises four sequential stages: (1) HAL co-occurrence matrix construction, (2) dimensionality reduction, (3) temperature-scaled attention pooling, and (4) classification.

\subsection{HAL Co-occurrence Matrix Construction}
Let $\mathcal{V}$ denote the vocabulary of size $V = |\mathcal{V}|$. To capture global word associations, we construct a co-occurrence matrix by passing a sliding window of size $W$ over the training corpus. For a given target word $w_i$ and a context word $w_j$ separated by a distance $d = |i - j|$, we assign a co-occurrence weight inversely proportional to their distance:
\begin{equation}
    f(w_i, w_j) = 
    \begin{cases} 
      \frac{1}{d} & \text{if } 0 < d \le W \\
      0 & \text{otherwise}
   \end{cases}
\end{equation}
Directionality is preserved by maintaining two separate sparse matrices: a left co-occurrence matrix $L \in \mathbb{R}^{V \times V}$ and a right co-occurrence matrix $R \in \mathbb{R}^{V \times V}$. The raw representation for a word $w \in \mathcal{V}$ is obtained by concatenating its corresponding row vectors from both matrices, resulting in a high-dimensional vector $v_{raw} \in \mathbb{R}^{2V}$.

\subsection{Dimensionality Reduction via Truncated SVD}
The raw matrix $M_{raw} = [L \oplus R] \in \mathbb{R}^{V \times 2V}$ exhibits extreme sparsity and a dimensionality that scales linearly with the vocabulary size. Utilizing $M_{raw}$ directly in a neural network induces severe memory constraints and increases the risk of overfitting. To mitigate this, we apply Truncated Singular Value Decomposition (SVD) to project $M_{raw}$ into a dense, continuous latent space:
\begin{equation}
    M_{raw} \approx U_k \Sigma_k V_k^T
\end{equation}
where $k \ll 2V$ is the target embedding dimension. Each word $w$ is subsequently mapped to a dense vector $x \in \mathbb{R}^k$. In our experiments, we set $W = 5$ and $k = 300$. These fixed embeddings are utilized as inputs to the subsequent neural layers.

\subsection{Temperature-Scaled Attention Pooling}
Given a sequence of length $T$, let $X = (x_1, x_2, \dots, x_T)$ represent the sequence of dense HAL embeddings. Standard mean pooling aggregates $X$ via an unweighted average, $s = \frac{1}{T} \sum_{t=1}^{T} x_t$, assigning equal importance to all tokens.

To extract contextually salient features, we introduce a parameterized additive attention layer. For each token embedding $x_t$, a scalar attention score $e_t$ is computed via a single-layer feed-forward network:
\begin{equation}
    e_t = v_a^\top \tanh(W_a x_t + b_a)
\end{equation}
where $W_a \in \mathbb{R}^{d_a \times k}$, 
$b_a \in \mathbb{R}^{d_a}$, and 
$v_a \in \mathbb{R}^{d_a}$ are learnable parameters.

Preliminary experiments indicated that standard softmax normalization often results in extreme weight concentrations on single tokens, leading to poor generalization. To explicitly regularize the attention distribution, we introduce a temperature scaling hyperparameter $\tau > 1$. The normalized attention weight $\alpha_t$ for each token is computed as:
\begin{equation}
    \alpha_t = \frac{\exp(e_t / \tau)}{\sum_{j=1}^{T} \exp(e_j / \tau)}
\end{equation}
The final sequence representation $s \in \mathbb{R}^k$ is the attention-weighted sum of the constituent token embeddings:
\begin{equation}
    s = \sum_{t=1}^{T} \alpha_t x_t
\end{equation}

\subsection{Classification and Regularization}
The aggregated vector $s$ is passed through a Multi-Layer Perceptron (MLP) for final classification. To ensure training stability and further prevent overfitting, we apply Layer Normalization \cite{ba2016layer} and Dropout \cite{srivastava2014dropout} prior to the final linear projection:
\begin{equation}
    h = \text{Dropout}(\text{ReLU}(\text{LayerNorm}(W_c s + b_c)), p)
\end{equation}
\begin{equation}
    \hat{y} = \text{Softmax}(W_o h + b_o)
\end{equation}
where $W_c, b_c, W_o, b_o$ are trainable parameters, and $p$ denotes the dropout probability. The entire architecture (comprising the attention mechanism and the classifier) is optimized end-to-end minimizing the standard cross-entropy loss function, augmented with $L_2$ weight decay regularization.
\section{Experiments and Results}

\subsection{Experimental Setup}
We evaluate the proposed architecture on the IMDB binary sentiment classification dataset \cite{maas2011learning}, comprising 25,000 training samples and 25,000 testing samples. Text preprocessing included lowercasing, HTML tag removal, and restricting the vocabulary to the $V = 10,000$ most frequent tokens. Sequences were truncated or padded to a fixed length of $T = 200$ tokens.

For the HAL matrix construction, we defined a context window of $W = 5$. The raw co-occurrence matrix was compressed to $k = 300$ dimensions using Truncated SVD. The fixed embeddings were fed into both the baseline and the proposed model. The baseline model employs standard mean pooling followed by the classifier, while the proposed model utilizes the temperature-scaled attention layer ($\tau = 2.0$) prior to the identical classifier. Both networks were trained using the Adam optimizer with a learning rate of $5 \times 10^{-4}$, a batch size of 64, and an $L_2$ weight decay of $1 \times 10^{-4}$. To prevent overfitting, early stopping was implemented with a patience of 5 epochs based on validation accuracy. Classifier dropout was set to $p = 0.6$.

\subsection{Quantitative Results}
Table \ref{tab:results} summarizes the classification performance. The traditional HAL representation combined with mean pooling achieves a test accuracy of 75.64\%. In contrast, integrating the parameterized attention mechanism yields a peak test accuracy of 82.38\%. This constitutes an absolute improvement of 6.74 percentage points under identical classifier configurations and embedding dimensions.

\begin{table}[h]
\centering
\caption{Comparative performance metrics: Initial (Epoch 1) vs. Peak accuracy on the IMDB test set.}
\label{tab:results}
\begin{tabular}{lccc}
\hline
\textbf{Model Architecture} & \textbf{Pooling Strategy} & \textbf{Initial Acc. (E1)} & \textbf{Peak Test Acc.} \\ \hline
Traditional HAL (1996) & Mean Pooling & 63.83\% & 75.64\% \\
\textbf{HAL-Attention (Proposed)} & \textbf{Attention Pooling} & \textbf{78.70\%} & \textbf{82.38\%} \\ \hline
\end{tabular}
\end{table}

The learning dynamics further emphasize the efficacy of the proposed mechanism. As illustrated in Figure 1, the attention-augmented model demonstrates rapid convergence, reaching 78.70\% accuracy within the first epoch. Conversely, the mean pooling baseline converges slowly and plateaus, primarily due to the persistent noise introduced by structural stop-words during the unweighted aggregation phase. 

\subsection{Qualitative Analysis and Interpretability}
Beyond performance metrics, attention-based pooling affords model interpretability. Mean pooling obscures the relative contribution of individual tokens. The proposed attention layer, however, outputs a normalized probability distribution over the sequence, quantifying token salience.

To validate this behavior, we examined a syntactically complex, mixed-sentiment sentence: \textit{"the cinematography was brilliant but the acting was completely awful and ruined the experience"}. 
A standard pooling mechanism would typically aggregate the positive token (\textit{"brilliant"}) and negative tokens (\textit{"awful", "ruined"}), risking a neutral or misclassified vector.

Our regularized attention mechanism successfully isolates the discriminative features and weighs the contrasting sentiments. The model assigned high attention weights to both the primary positive term, \textit{brilliant} ($\alpha = 0.2960$), and the primary negative modifiers, \textit{awful} ($\alpha = 0.2662$) and \textit{completely} ($\alpha = 0.2354$). The temperature scaling factor ($\tau = 2.0$) prevented the distribution from collapsing onto a single trigger word, allowing the model to capture the competing sentiments before correctly classifying the document as negative. This confirms that the proposed architecture not only filters out structural noise but also mimics a more holistic, context-aware reading strategy.

 \begin{figure}[h]
     \centering
     \includegraphics[width=0.8\textwidth]{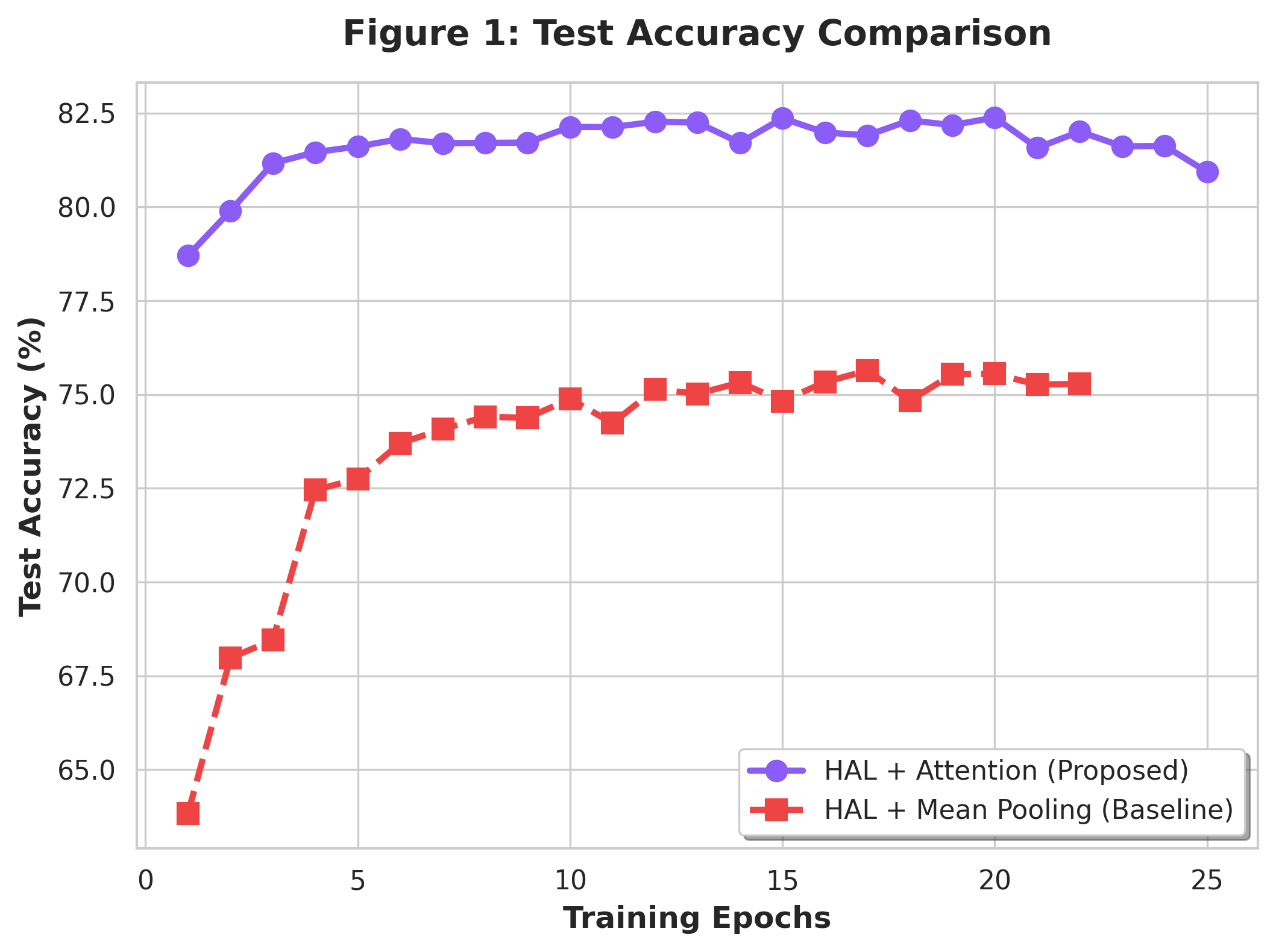}
     \caption{Test accuracy convergence over training epochs.}
     \label{fig:accuracy}
 \end{figure}

\section{Conclusion}

This paper presented an enhanced architecture for aggregating distributional semantic representations derived from the Hyperspace Analogue to Language (HAL) model. By replacing the conventional mean pooling strategy with a parameterized, temperature-scaled attention mechanism, we addressed the inherent information loss associated with unweighted sequence aggregation. The application of Truncated SVD effectively mitigated the extreme dimensionality and sparsity of the raw co-occurrence matrices, enabling efficient end-to-end neural network training. 

Empirical evaluation on the IMDB sentiment analysis dataset demonstrated that the proposed attention-based pooling mechanism yields an absolute accuracy improvement of 6.74 percentage points over the traditional baseline. Furthermore, qualitative analysis confirmed that the attention layer provides an interpretable aggregation scheme, successfully filtering out uninformative structural tokens while dynamically weighting context-salient features, even in semantically complex sequences.

Future work will investigate the scalability of this attention-augmented distributional framework across diverse natural language processing tasks, including document categorization and natural language inference. Additionally, we aim to explore the integration of this mechanism with sub-word tokenization strategies to handle out-of-vocabulary terms in highly specialized corpora.

\section*{Declarations}

\noindent \textbf{Author Contributions:} \textbf{A. Sakour} conceptualized the proposed attention-based pooling architecture, developed the codebase, conducted the experiments, and served as the primary author of the manuscript. \textbf{Z. Sakour} assisted in refining the methodology, performing code review, and proofreading the manuscript. 

\vspace{1em}
\noindent \textbf{Declaration of AI Usage:} During the preparation of this manuscript and the associated codebase, the authors utilized Large Language Models (LLMs) to assist with code optimization, the generation of data visualizations, and the refinement of academic English prose. Following the use of these tools, the authors rigorously reviewed, validated, and edited all AI-assisted outputs, taking full responsibility for the final content and integrity of the research.

\end{document}